\pdfoutput=1

\documentclass[runningheads]{llncs}
\usepackage{graphicx}

\usepackage{tikz}
\usepackage{comment}
\usepackage{amsmath,amssymb} 
\usepackage{color}

\usepackage[width=122mm,left=12mm,paperwidth=146mm,height=193mm,top=12mm,paperheight=217mm]{geometry}

\begin{document}
\pagestyle{headings}
\mainmatter
\def\ECCVSubNumber{003}  

\title{GCF-Net: Gated Clip Fusion Network for Video Action Recognition} 

\titlerunning{GCF-Net}
%
\author{Jenhao Hsiao \and
Jiawei Chen \and
Chiuman Ho}
\authorrunning{J.H. Hsiao et al.}
%
\institute{InnoPeak Technology, Palo Alto, CA, USA
\email{\{mark,jiawei.chen,chiuman\}@innopeaktech.com}}
\maketitle

\begin{abstract}

In recent years, most of the accuracy gains for video action recognition have come from the newly designed CNN architectures (e.g., 3D-CNNs). These models are trained by applying a deep CNN on single clip of fixed temporal length. Since each video segment are processed by the 3D-CNN module separately, the corresponding clip descriptor is local and the inter-clip relationships are inherently implicit. Common method that directly averages the clip-level outputs as a video-level prediction is prone to fail due to the lack of mechanism that can extract and integrate relevant information to represent the video. 

In this paper, we introduce the Gated Clip Fusion Network (GCF-Net) that can greatly boost the existing video action classifiers with the cost of a tiny computation overhead. The GCF-Net explicitly models the inter-dependencies between video clips to strengthen the receptive field of local clip descriptors. Furthermore, the importance of each clip to an action event is calculated and a relevant subset of clips is selected accordingly for a video-level analysis. On a large benchmark dataset (Kinetics-600), the proposed GCF-Net elevates the accuracy of existing action classifiers by 11.49\% (based on central clip) and 3.67\% (based on densely sampled clips) respectively.
\end{abstract}

\section{Introduction}

\begin{figure*}
\begin{center}
\includegraphics[width=0.9\linewidth]{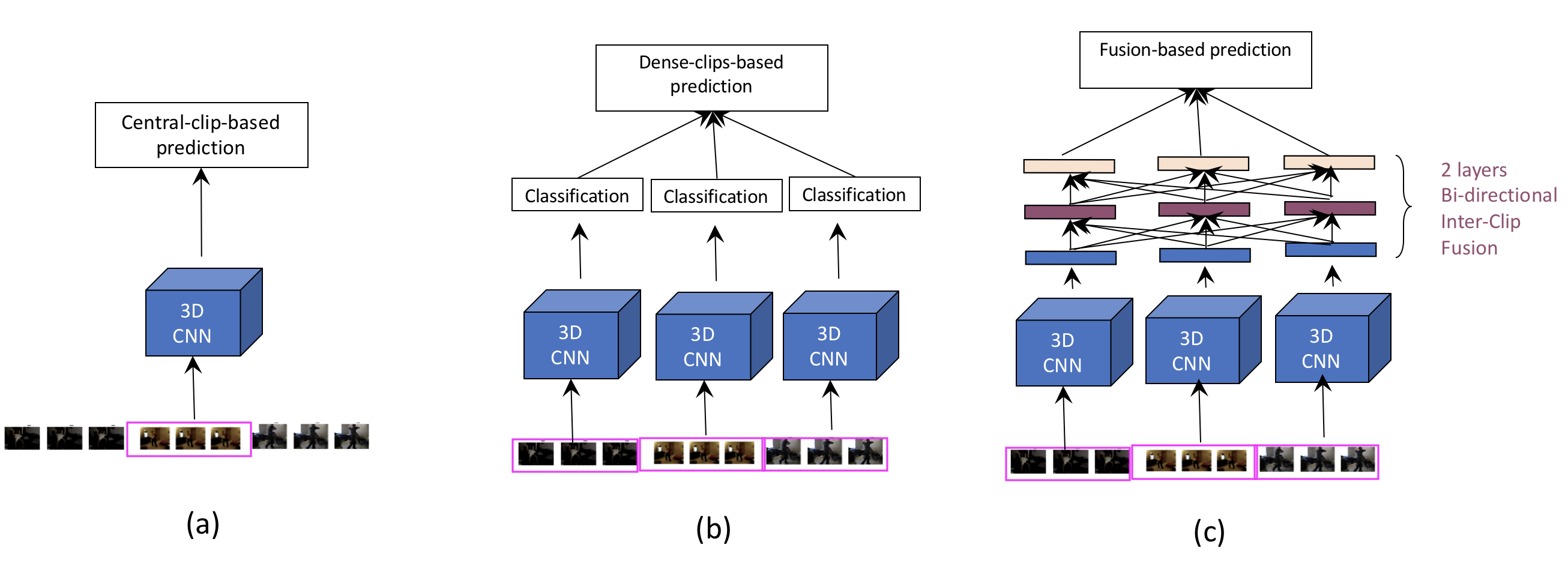}
\end{center}
    
   \caption{Methods of video-level prediction: (a) central clip prediction; (b) dense clips prediction; (c) our bi-directional inter-clip fusion based prediction}
\label{fig:videopred}
\end{figure*}

The explosive growth in video and its applications has drawn considerable interest in the computer vision community and boost the need of high-level video understanding. Action recognition and localization in videos are the key research problems for high-level video understanding, where action recognition is to classify a video by assigning a pre-defined action classes, while action localization determines whether a video contains specific actions and also identifies temporal boundaries (e.g., start time and end time) of each action instance.

Video action recognition has witnessed much good progress in recent years. Most of the accuracy gains have come from the introduction of new powerful 3D-CNN architectures \cite{Carreira2017QuoVA,tranvcvpr18,qiuiccv17,Tran15learningspatiotemporal,Taylor_convolutionallearning,DBLP:journals/corr/abs-1811-08383}. However, there are some drawbacks and limitations of these 3D-CNN based architectures. Firstly, most modern action recognition models are trained by applying a deep CNN on single clip of fixed temporal length (e.g., commonly 16 or 32 frames as a clip in existing works). This will become a performance bottleneck since the duration of different actions are variant and complex actions span multiple video segments (i.e., multiple clips can be involved). Single clip captures only very limited local temporal knowledge (e.g., may be either relevant or irrelevant to the target action), and it can hardly describe an action accurately from a global view. It is thus prone to fail due to the challenges in extracting global information.

Secondly, to achieve a video-level action prediction, most existing architectures adopt the following naive methods: central clip prediction or dense clip prediction. Central clip prediction (as shown in Figure \ref{fig:videopred}(a)) directly assume that the central clips (e.g., the central one or few clips) are the most related event, and average the action predictions of those clips to deliver a video-level action prediction. Dense clips prediction (as shown in Figure \ref{fig:videopred}(b)), on the other hand, aggregates the clip-level predictions over the entire video via a similar averaging scheme to model the temporal structure. Although these naive methods are easy to implement, they could somewhat hurt the action recognition accuracy. Videos in the real-world actually exhibit very different properties: they are often several minutes long, where brief relevant clips are often interleaved with segments of extended duration containing irrelevant information. Directly pooling information from all clips (or a few central clips) without consideration of their relevance will cause poor video-level prediction, as irrelevant clips may dominate the decision process.

Lastly, to detect an action's temporal location in a video, a strong supervision, in the form of training videos that have been manually collected, labeled and annotated, will be heavily relied on. However, collecting large amounts of accurately annotated action videos is already very expensive. Acquiring a large amount of labeled video data with temporal boundary information will be prohibitively difficult.

In this paper, we introduce the Gated Clip Fusion Network (GCF-Net) that uses a more sophisticated method to extract and integrate the information from each clip, and greatly boost the existing video action classifier with the cost of a tiny computation overhead. The proposed GCF-Net explicitly models the inter-dependencies between video clips and captures the clip-wise importance to increase the deep network’s sensitivity to informative features across video segments, which turn out to deliver a more accurate recognition due to the use of a more comprehensive video-level feature.  As a by-product, the enhanced prediction can be unsupervisedly back-propagated to generate an estimated spatio-temporal map that localizes possible actions in videos. An overview of our algorithm is shown in Figure \ref{fig:GCF-Net}.

The contributions of this paper are summarized as below.
\begin{itemize}
  \item We present a novel Bi-directional Inter-Clip Fusion method (Figure \ref{fig:videopred}(c)) that is able to utilize both short- and long-range video segments to model the inter-clip relationships and generate better clip representations. Comparing to traditional methods that mainly rely on 3D-CNN to separately generate a local feature with very limited local temporal knowledge, our method provides a better clip representation with broader receptive field.
  \item A Gated Clip-Wise Attention is proposed as the means to further suppress irrelevant clips for improving the video-level prediction accuracy. As a by-product, the attention weights generated by this module can be used to locate the time interval of an action event in a video (e.g., based on relevant clips).
  \item We demonstrate that the proposed GCF-Net, which models the inter-clip relationships and clip-wise importance in a much finer granularity, yields  significant gain in video action recognition accuracy comparing to traditional methods that conduct analysis on all clips (e.g., dense sampling) or randomly/centrally selected clips. On a large benchmark video dataset (Kinetics-600), our method elevates the accuracy of an already state-of-the-art action classifier by by 11.49\% (based on central clip) and 3.67\% (based on densely clips sampling) respectively with the same amount of training data set and backbone network.

\end{itemize}
The rest of this paper is organized as follows. We discuss the related work in Section 2 and describe our action GCF-Net in Section 3. Section 4 presents the details of our experiment and Section 5 concludes this paper.

\section{Related work}

The goal of video action recognition aims to identify a single or multiple actions per video, while action localization further attempts to determine the action intervals in a video. In recent years, most of the accuracy gains for video action recognition have come from the introduction of new powerful architectures. 

Before the success of CNNs, hand-designed video features \cite{Laptev03space-timeinterest,piotrconf,Sadan_actionbank:} was the mainstream approach and methods on improved dense trajectories \cite{Wang13actionrecognition} presented good performance. When it comes to the era of deep learning, convolutional neural networks have delivered a major paradigm shift and have been widely used to learn video features and classify video in an end-to-end manner. Two-stream method \cite{twostream} is one of the popular frameworks that integrates spatial and temporal information via 2D network. 3D ConvNets \cite{Tran15learningspatiotemporal,Taylor_convolutionallearning} extend aforementioned 2D image models to the spatio-temporal domain, handling both spatial and temporal dimensions in a similar convolution way. A combination of two-stream networks and 3D convolutions, known as I3D \cite{Carreira2017QuoVA}, was proposed as a generic video representation learning method. There are also methods focusing on decomposing the convolutions into separate 2D spatial and 1D temporal filters \cite{tranvcvpr18,qiuiccv17}, while methods in \cite{9008780,cvpr-wang-18,pami18-varol} model long-term filtering and pooling using different strategies (e.g., temporal strides, slow and fast network, self-attention, and etc.). In \cite{DBLP:journals/corr/abs-1811-08383}, the authors try to achieve the performance of 3D CNN but maintain 2D CNN’s complexity. 

For action localization, the objective is to localize the temporal start and end of each action within a given untrimmed video and to recognize the action class. Most of the existing methods \cite{sst17,humam18,fabian16,tianwei18} hugely rely on large-scale annotated video data, and achieve the detection goal through a two-step mechanism: an action proposal method firstly identifies candidate action segments, and then a classifier validates the class of each candidate and refines its temporal boundaries.

However, the networks proposed by the above methods mainly focused on the design of convolution network architecture, and are trained by single clip (e.g., gradients are updated based on one clip point of view), where irrelevant video segments could lead the gradient to the wrong direction and thus disrupt training performance. During inference, clips are sampled densely or at random. This naive averaging strategy can hardly model a complex action event that spans multiple video segments, and will hurt the recognition accuracy since, again, irrelevant clips could negatively affect the prediction decision.

Another branch of methods is the so-called learning-to-skip approach \cite{Hehe18,ZuxuanWu19,serena16}, where they skip segments by leveraging past observations to predict which future frames to consider next. In \cite{Korbar_2019_ICCV}, the authors proposed a salient clip sampling strategy that prevents uninformative clips from joining the video-level prediction. However, the above methods merely skip or sample clips in videos, and didn't fully utilize the inter-clip relationships to model the temporal relationship of an action. The accuracy improvement can thus be limited.

\section{Method}

\begin{figure*}
\begin{center}
\includegraphics[width=0.9\linewidth]{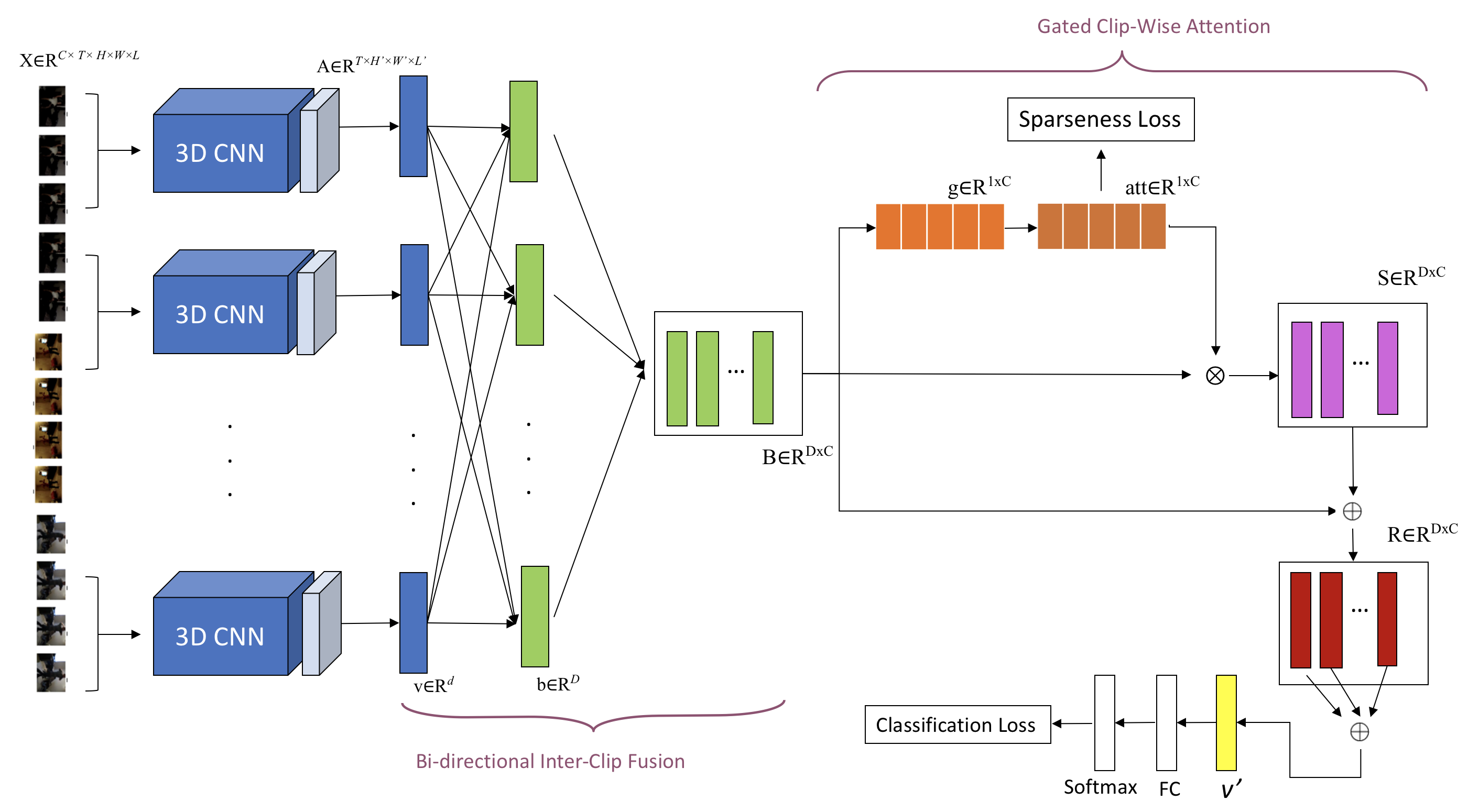}
\end{center}
    
   \caption{The proposed GCF-Net.}
\label{fig:GCF-Net}
\end{figure*}

We claim that a successful action recognition network should consider both the inter-clip dependencies and clip-wise importance. Inter-clip dependencies expand the view of local clip descriptor to model a complex and multi-segment action event, while clip-wise importance identifies a set of key segments presenting important action component so that irrelevant clips can be prevented from joining the video-level prediction. To achieve this goal, our proposed GCF-Net, which is an end-to-end deep neural network, introduces two key modules: Bi-directional Inter-Clip Fusion module and Gated Clip-Wise Attention module. We describe each step of the proposed GCF-Net in the rest of this section.

\subsection{Gated Clip Fusion Network}

\subsubsection{Backbone Network}
Figure \ref{fig:GCF-Net} shows an overview of the proposed framework. For the proposed framework, clips $X = \{x_1, x_2, …, x_C\}$ are set as the inputs and each clip $x$ contains $T$ stacked frames. Each input clip will be firstly processed by 3D-CNN, which contain a set of 3D convolutional layers, to extract corresponding clip features. The input shape for one batch data is $C \times T \times H \times W \times L$, where $C$ denotes the number of clips, $T$ frames are stacked together with height $H$ and width $W$, and the channel number $L$ is 3 for RGB images. The convolutional kernel for each 3D convolutional layer in 3D-CNN is in 3 dimensions. Then for each 3D convolutional layer, data will be computed among the three dimensions simultaneously.

To maintain the temporal accuracy of action localization, no strides will be applied along the T dimension during the 3D convolution. The feature map in the last convolution layer, denoted as A, is thus having dimension of $T \times H' \times W' \times L'$, where $H’$ and $W’$ are the height and width of feature maps, and $L’$ is the number of convolution filters in the last convolution layer. Global pooling is then used to pool the content of feature maps and generate a summary vector of single clip. Outputs of 3D-CNN step are thus a set of raw clip descriptors $V = \{v_1, v_2, …, v_C\}$, where $v \in R^d$ is the output of global pooling layer in 3D-CNN. 

\subsubsection{Bi-directional Inter-Clip Fusion}

Since each clip descriptor is produced by the 3D-CNN module separately, the inter-clip relationships modeled by convolution are inherently implicit and local. That is, each clip descriptor can only observe an extremely limited local event and there are no inter-clip relationships. This will become a performance bottleneck since the duration of different actions are variant and complex actions could span multiple video segments (e.g., multiple clips can be involved for an action event). 

To capture the inter-clip dependencies for both short- and long-range dependencies, we propose the Bi-directional Inter-Clip Fusion that aims at strengthening the local clip descriptor of the target position via aggregating information from other positions (e.g., other video segments). Motivated by \cite{NIPS2017_7181}, the inter-clip relationships can be fused by a bi-directional attention to link different clips and can be expressed as:
\begin{equation}
    BA(v_i)=W_Z\sum_j\frac{(W_qv_i)(W_kv_j^T)}{N(v)} (W_vv_j),
\end{equation}
where $i$ is the index of the target positions, and $j$ enumerates all possible other clip positions. $W_q$, $W_k$, $W_v$ and $W_z$ denote linear transform matrices. $(W_qv_i)(W_kv_j)$ denotes the relationship between video clip $i$ and $j$, and $N(v)$ is the normalization factor. The resulting fused clip descriptor is named as bi-directional clip descriptor $b_i = BA(v_i)$, and $B=\{b_1, b_2, ..., b_C\}$, where $b \in R^D$ is a $D$-dimensional vector.

\subsubsection{Gated Clip-Wise Attention}

An action typically represents only a subset of objects and events which are most relevant to the context of a video. To suppress irrelevant clips, we further introduce the clip-wise attention to re-weight the above bi-directional clip descriptors and remove less related clips to join the final video-level action recognition decision. With a designated loss function (details described in next section), the resulting attention vector can also help locate temporal intervals of an action unsupervisely. 

To be more specific, we first generate clip-wise statistics for each bi-directional clip descriptor by global average pooling
\begin{equation}
g = [mean(b_1), mean(b_2), ..., mean(b_{C})]
\end{equation}
The pooled output can be interpreted as a summary of the bi-directional clip descriptor whose statistics are expressive for the whole clip. To fully capture clip-wise dependencies, here we employ a gating mechanism with a sigmoid activation:
\begin{equation}
att=\sigma_{sigmoid}(W_2\sigma_{ReLU}(W_1g)),
\end{equation}
where $\sigma_{ReLU}$ refers to the ReLU function, $W_1$ and $W_2$ are the fully connected layer weights, and $\sigma_{sigmoid}$ is the sigmoid function. The $att$ weight vector is defined in a clip-agnostic way, which is useful to identify video segments that are relevant to the action of interest and estimate the temporal intervals of the detected actions.

The final output of the block is obtained by re-scaling clip descriptors with the activation $att$:

\begin{equation}
S = \{s_1, s_2, …, s_C\},
\end{equation}
\begin{equation}
s_i = att_i \times b_i
\end{equation}

The $att$ introduces dynamics conditioned on the input (i.e., bi-directional clip descriptor), which can be regarded as a gated function that re-weights the clips based on their significance to an action event. 

To make the learning more robust and effective, a residual module is introduced 
\begin{equation}
R = B + S 	
\end{equation}
where $R = \{r_1, r_2, …, r_c\}$ ($r \in R^D$) can be considered as the residual clip descriptors. Finally, a video-level representation, denoted by $v’$, corresponds to the residual clip descriptors R, is given by
\begin{equation}
v'=\frac{\sum_ir_i}{C}
\end{equation}
The action recognition can be performed based on $v’$ and is defined as
\begin{equation}
y = \sigma_{softmax}(W_3v’),
\end{equation}
where $W_3$ is the fully connected layer weights, and $\sigma_{softmax}$ is the softmax function.

\subsection{Loss Function}
The loss function in the proposed network is composed of two terms, the action classification loss and the sparsity loss, which is given by
\begin{equation}
L = L_c + \lambda L_s
\end{equation}
where $L_c$ denotes the classification loss computed on the video-level action labels, $L_s$ is the sparsity loss on the clip-wise attention weights, and $\lambda$ is a constant to control the trade-off between the two terms. The classification loss is based on the standard cross-entropy loss between ground truth and the prediction y (generated by GCF-Net), while the sparsity loss is given by the $L_1$ norm on attention weights $att$:
\begin{equation}
L_s = \parallel att \parallel_1
\end{equation}
The output of attention weights will have the tendency towards 0 or 1 due to the use of $L_1$ loss and sigmoid function. In this case, action-related clips can be recognized with a sparse subset of key segments in a video, which will help locating the relevant clips for action detection.

\subsection{Training Details}

For the training of GCF-Net on Kinetics-600, Stochastic Gradient Descent (SGD) with standard categorical cross entropy loss is applied, and we use 128 videos as mini-batch size of SGD. The momentum, dampening and weight decay are set to 0.9, 0.9 and $1\times10^{-3}$, respectively. Learning rate is initialized with 0.1 and reduced 3 times with a factor of $10^{-1}$ when the validation loss converges. Note that the network was trained from scratch and no other dataset is used as prior. For the training of UCF-101 benchmark, we have used the pretrained models of Kinetics-600. We have frozen the network parameters and fine-tuned only the last layer. For fine-tuning, we start with a learning rate of 0.01 and reduce it with a factor of $10^{-1}$ when the validation loss converges. For spatial augmentation, we perform multi-scale cropping to augment video data. For Kinetics-600 and UCF-101, input clips are flipped with 50\% probability.

For action recognition inference on Kinetics-600 and UCF-101, we select non-overlapping 16-frame from each video sample as one clip, and 10 clips are used as the network input. If the video contains smaller number of clips than
the input size, loop padding is applied. After the padding, input to the network has the size of $10 \times 16 \times 112 \times 112 \times 3$ referring to number of clip numbers, frames, width, height, and input channels respectively.

\section{Experiments}

In this section, we first explain the experimented datasets. Then, we discuss about the achieved results for the experimented network architectures.

\subsection{Datasets}

\begin{itemize}
  \item Kinetics-600 dataset \cite{kinetics600} contains 600 human action classes, with at
least 600 video clips for each action. Each clip is approximately 10 seconds long and is taken from a different YouTube video. There are in total 392,622 training videos. For each class, there are also 50 and 100 validation and test videos,

  \item UCF101 dataset \cite{soomro2012ucf101} is an action recognition dataset of realistic action videos, collected from YouTube. It consists of 101 action classes, over 13k clips and 27 hours of video data. Compared to Kinetics-600, UCF-101 contains very little amount of training videos, hence prone to over-fitting. We thus conduct transfer learning (from network learned by Kinetics-600) to avoid this problem and show the effectiveness of the proposed method.
\end{itemize}

\subsection{Action Recognition Accuracy}

\begin{table}
\begin{center}
\begin{tabular}{|l|c|c|}
\hline

Method & \multicolumn{2}{|c|}{Top1 Accuracy(\%)} \\
& Kinetics-600 & UCF101 \\
\hline\hline
3D-MobileNet+C-Clip  & 42.79 &  60.29 \\
3D-ShuffleNet+C-Clip  & 45.61 & 61.27 \\
R(2+1)D+C-Clip  & 54.69 & 76.92\\
3D-ResNeXt-101+C-Clip  & 58.58 &  80.12\\
\hline\hline
3D-MobileNet+D-Clip  & 48.35 &  71.60 \\
3D-ShuffleNet+D-Clip  & 53.70 & 73.32 \\
R(2+1)D-101+D-Clips  & 62.18 & 87.02 \\
3D-ResNeXt-101+D-Clips  & 66.40 & 89.08 \\
\hline\hline
GCF-Net(3D-MobileNet) & 52.53 & 80.32\\
GCF-Net(3D-ShuffleNet) & 57.10 & 81.23\\
GCF-Net(R(2+1)D) & 68.01 & 95.12 \\
GCF-Net(3D-ResNeXt-101) & 70.07 & 96.82\\

\hline
\end{tabular}
\end{center}
\caption{Accuracy comparison of different methods}
\label{tab:recog-accuracy}
\end{table}

In this subsection, we study the effectiveness of the proposed model on learning video representations on different dataset.

Our GCF-Net can be used with any clip-based action classifiers and immediately boost the recognition accuracy. We demonstrate the general applicability of our approach by evaluating it with several popular 3D CNNs. For R(2+1)D network \cite{tranvcvpr18}, we implement it by replacing the original 3D convolutional kernels (e.g.,$ t \times k \times k$) with two  (virtually) 2D blocks (e.g., $1 \times k \times k$  and $t \times 1 \times 1$ kernels) in 3D-ResNet-101 (i.e., 3D-ResNet with 101 layers). 3D-ResNext-101 \cite{8578783} \cite{8100117} is an more efficient version of 3D-ResNet due to the use of cardinality block, and it also has 101 layers in our experiments. In addition to the above full networks, we also implement light weight networks for performance comparison, including 3D-MobileNet \cite{Sandler_2018_CVPR} and 3D-ShuffleNet \cite{Ma_2018_ECCV}. 3D-MobileNet and 3D-ShuffleNet are the extentions of their original 2D versions (i.e., adding temporal dimension into kernel).

Table \ref{tab:recog-accuracy} shows the action recognition results of all methods. The baseline for our evaluation is the central-clip-based prediction (noted as C-Clip in the Table). For Kinetics-600 dataset, as can be seen that the baseline central-clip method has the poorest top-1 accuracy. Full networks, such as R(2+1)D and 3D-ResNeXt-101, can only achieves 54.69\% and 58.58\% top-1 accuracy, while light weight network, such as 3D-MobileNet and 3D-ShuffleNet, delivers even poorer 42.79\% and 45.61\% top-1 accuracy. The poor performance is mainly due to the lack of fully utilizing the information in the video (e.g., the rest relevant clips). Another popular method, dense-clips-based prediction (noted as D-Clips in the table) achieves better recognition accuracy (comparing to central-clip prediction). Among all dense-clips-based methods, 3D-ResNeXt-101 (i.e., 3D-ResNeXt-101+D-Clips) achieves the best 66.40\% top-1 accuracy. 

However, since an action is usually complex and spans video segments, uniformly averaging all clips is obviously not the best strategy and can only achieve limited accuracy. As can be seen in Table \ref{tab:recog-accuracy}, our GCF-Net outperforms all the central-clip and dense-clips-based methods on the same backbone networks. The best top-1 accuracy, delivered by GCF-Net based on 3D-ResNeXt-101, achieves 70.07\% top-1 accuracy, which outperforms 3D-ResNeXt-101+D-Clips and 3D-ResNeXt-101+C-Clip  by 3.67\% and 11.49\% respectively. The improvement on light weight networks are more significant (in terms of relative percentage of change), where GCF-Net(3D-ShuffleNet) outperforms 3D-ShuffleNet+D-Clip and 3D-ShuffleNet+C-Clip by 3.4\% (i.e., 6.3\% boosting) and 11.49\% (i.e., 25.2\% boosting) respectively. The proposed GCF-Net is thus proved to be able to build a better video-level feature representation that can capture short- and long-range statistics, which delivers a significant better recognition accuracy.

For UCF-101 dataset, we observed similar performance improvement. Comparing to traditional methods, such as 3D-ResNeXt-101+D-Clips and 3D-ResNeXt-101+C-Clip, that only have 80.12\% and 89.08\% accuracy, our GCF-Net(3D-ResNeXt-101) delivers a significant boosting in accuracy, which is 96.82\%.

\subsection{Ablation Studies}

\begin{table}
\begin{center}
\begin{tabular}{|l|c|}
\hline
Method & Accuracy(\%) \\
\hline\hline

C-Clip prediction & 58.58\\
D-Clips prediction & 66.40\\
\hline\hline
Inter-Clip Fusion only & 69.01\\
Clip-Wise Attention only & 68.46\\
GCF-Net Full & 70.07\\

\hline
\end{tabular}
\end{center}
\caption{Accuracy comparison of different GCF-Net modules on Kinetics dataset (all methods are based on 3D-ResNeXt-101)}
\label{tab:fusion-exp}
\end{table}

In this subsection we further investigate the accuracy improvement brought by the individual inter-clip (i.e., Bi-directional Inter-Clip Fusion) and clip-wise (Gated Clip-Wise Attention) strategies.

To implement inter-clip-fusion-only strategy, the output of bi-directional clip descriptor $B$ is directly used to generate a video-level representation (i.e., $ v'=\frac{\sum_ib_i}{C} $), where the rest clip-wise attention module in the original GCF-Net is skipped. For clip-wise-attention-only strategy, the output of the raw 3D-CNN feature $V$ is directly fed into clip-wise attention module (i.e., $g = [mean(v_1), mean(v_2), ..., mean(v_{C})]$), and the same fusion process is proceeded to generate a video-level representation for later prediction. 

Table \ref{tab:fusion-exp} shows the accuracy comparison of different GCF-Net modules. As can be seen both strategies outperform traditional video-level prediction methods (e.g., C-Clip and D-Clips). Inter-clip-fusion-only strategy achieve slightly better top-1 accuracy of 69.07\% than clip-wise-attention-only strategy that has 68.46\% accuracy, which shows the importance of inter-clip dependencies. Since the raw clip descriptor generated by 3D-CNN can only observe an extremely limited local information, using the clues from neighbors will largely extend the receptive field and strengthen the clip descriptor. The inter-clip fusion thus shows its excellence in fusing the inter-clip relationships to better describe a video segment. 

Despite the fact that clip-wise-attention-only strategy's performance is a bit inferior to inter-clip-fusion-only strategy, it can be a complement to inter-clip strategy and can further refine the bi-directional clip feature (e.g., by filtering out irrelevant ones), which helps generating a better video-level representation. By concatenating clip-wise module with inter-clip, we can deliver the best video-level prediction accuracy, which proves the effectiveness of each module and the overall strategy.

\subsection{Feature Dimension and Fusion Layers}

\begin{figure}[t]
\begin{center}
\includegraphics[width=1.0\linewidth]{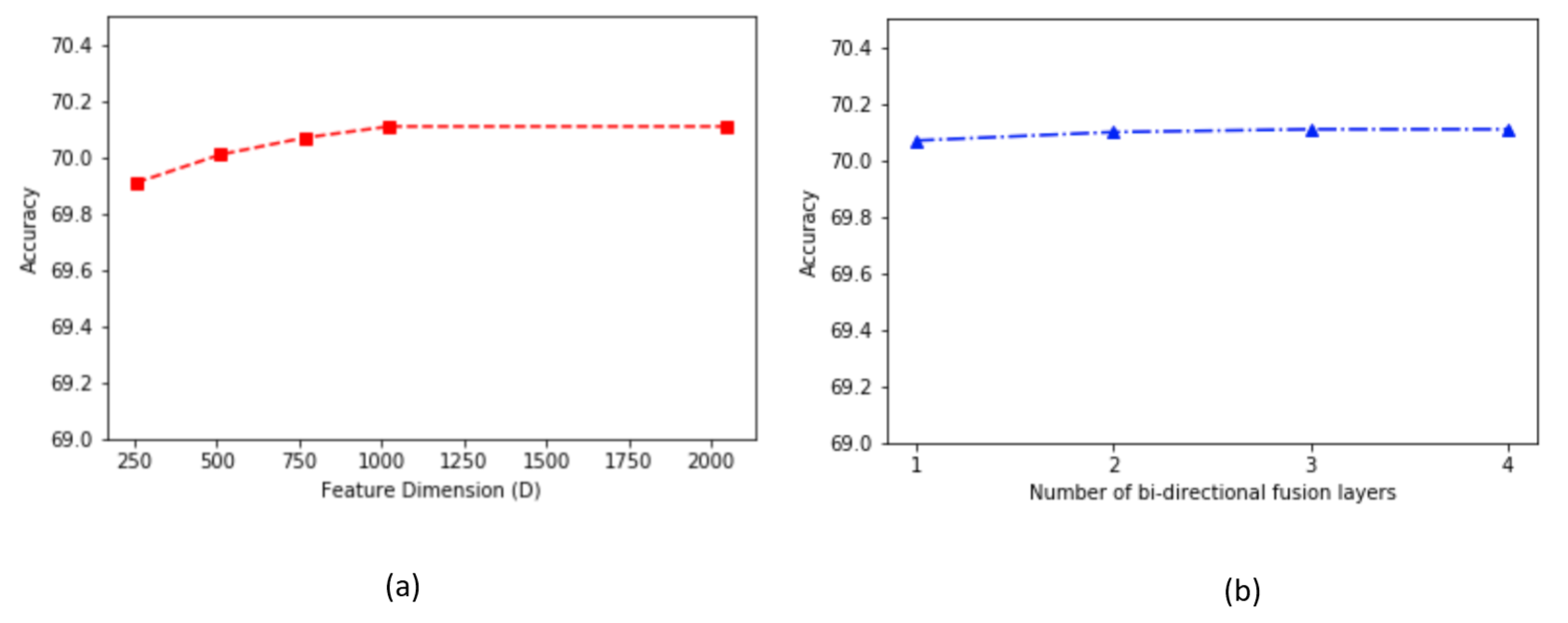}
\end{center}
   \caption{Performance comparison of Feature Dimension and Fusion Layers}
\label{fig:fusion-layers}
\end{figure}

In this subsection we explore the effect of network parameters in terms of prediction accuracy.

Figure \ref{fig:fusion-layers}(a) shows the result of different length of features, where $D$ is the parameters to determine the feature dimension of a clip descriptor $b$, residual clip descriptor $r$, and video-level descriptor $v'$. As can be seen the GCF-Net has good performance in generating compact representation of video-level descriptor. Comparing to the raw 3D-CNN pooled feature (e.g., 2048-d in 3D-RestNeXt), GCF-Net only needs as small as a 256-d vector to represent a video-level, which can still achieve better top-1 accuracy than the central- and dense-clips-based methods, and is 8 times smaller than the raw 3D-CNN feature. Increasing the size of feature dimension in GCF-Net can slightly improve the accuracy, but it soon gets saturated after $D >$ 1024.

For Bi-directional Inter-Clip Fusion, we can concatenate multiple bi-directional fusion layers to have a deeper fusion of inter-clip  dependencies:

\begin{equation}
BA^n=BA( BA^{n-1} )    
\end{equation}
\begin{equation}
BA^1=BA(v_i)
\end{equation}
where $n$ is the number of inter-clip fusion layers. Figure \ref{fig:fusion-layers}(b) show the result of different number of bi-directional fusion layers. As can be seen the increment of layers has less effect on the accuracy, which reveals that the GCF-Net can model the inter-clip relationship very effectively using jsut a few layers. It can achieve achieve $>$ 70\% top-1 accuracy when $n = 1$ and gains a bit more when the number of layers increases.  

\subsection{Complexity Level}


\begin{table}
\begin{center}
\begin{tabular}{|l|c|c|}
\hline

Method & MFLOPs & Params \\
\hline\hline
3D-MobileNet  & 446 &  3.12M \\
3D-ShuffleNet & 360 & 6.64M \\
R(2+1)D & 6,831 & 50.91M \\
3D-ResNeXt-101 & 6,932 & 48.34M \\
\hline\hline
GCF-Net(w/o backbone network) & 15 & 0.86M \\

\hline
\end{tabular}
\end{center}
\caption{Model complexity}
\label{tab:complexity}
\end{table}

Table \ref{tab:complexity} shows the complexity level of each model. As expected that the light weight networks (e.g., 3D-MobileNet and 3D-ShuffleNet) have much smaller MFLOPs and parameters than full network (e.g., 3D-ResNeXt and R(2+1)D). We can observe that, comparing to the backbone networks, the overhead brought by GCF-Net (without the backbone network) is actually quite tiny. It has 15 MFOPs and 0.86M network parameters, which occupies only 3.2\% and 0.2\% computation effort of the overall prediction pipeline for 3D-MobileNet- and 3D-ResNeXt-based frameworks. Hence, GCF-Net successfully upgrades the video-level prediction framework with the cost of a tiny MFLOPs increment in exchange of a significant boost of accuracy.

\subsection{Visualization}

To verify the ability of GCF-Net on locating action event in a video, here we extend the the 2D Grad-CAM \cite{Selvaraju_2017_ICCV} to a 3D spatio-temporal action localization map. Similar to \cite{Selvaraju_2017_ICCV}, we first compute the gradient of the predicted action class $m$ (i.e., $y^m$) with respect to feature maps $A_{tk}$ of the last convolutional layer at moment $t$ (e.g., the output of $k$-th convolution kernel for $t$-th frame in video). These flowing-back gradients are global average-pooled to obtain the neuron importance weights $\alpha_{t,k}$:
\begin{equation}
\alpha^m_{t,k}=\frac{1}{Z}\sum_i\sum_j\frac{\partial y^m}{\partial A^{t,k}_{ij}}
\end{equation}
where $Z$ is the normalization factor (i.e., the size of feature map). The weight $\alpha^m_{t,k}$ represents a partial linearization of the deep network downstream from $A^m_{t,k}$, and captures the ‘importance’ of feature map $k$ for a target action $m$. 

After having the importance estimation of each channel, we perform a weighted combination of forward activation maps, and follow it by a ReLU to obtain the localization map:
\begin{equation}
M_t=\sigma_{ReLU}(\sum_k\alpha^m_{t,k}A^{t,k})
\end{equation}
The output can then be resized to match the size of original video frames to showcase the spatial-temproal heatmap of an action event. Since only positive weights are what we interested (e.g., positive influence on the target action class), we can apply a ReLU to the linear combination of maps. In this case, we can get the intensity level of pixels whose intensity should be increased in order to increase $y^m$.

\begin{figure*}
\begin{center}
\includegraphics[width=0.9\linewidth]{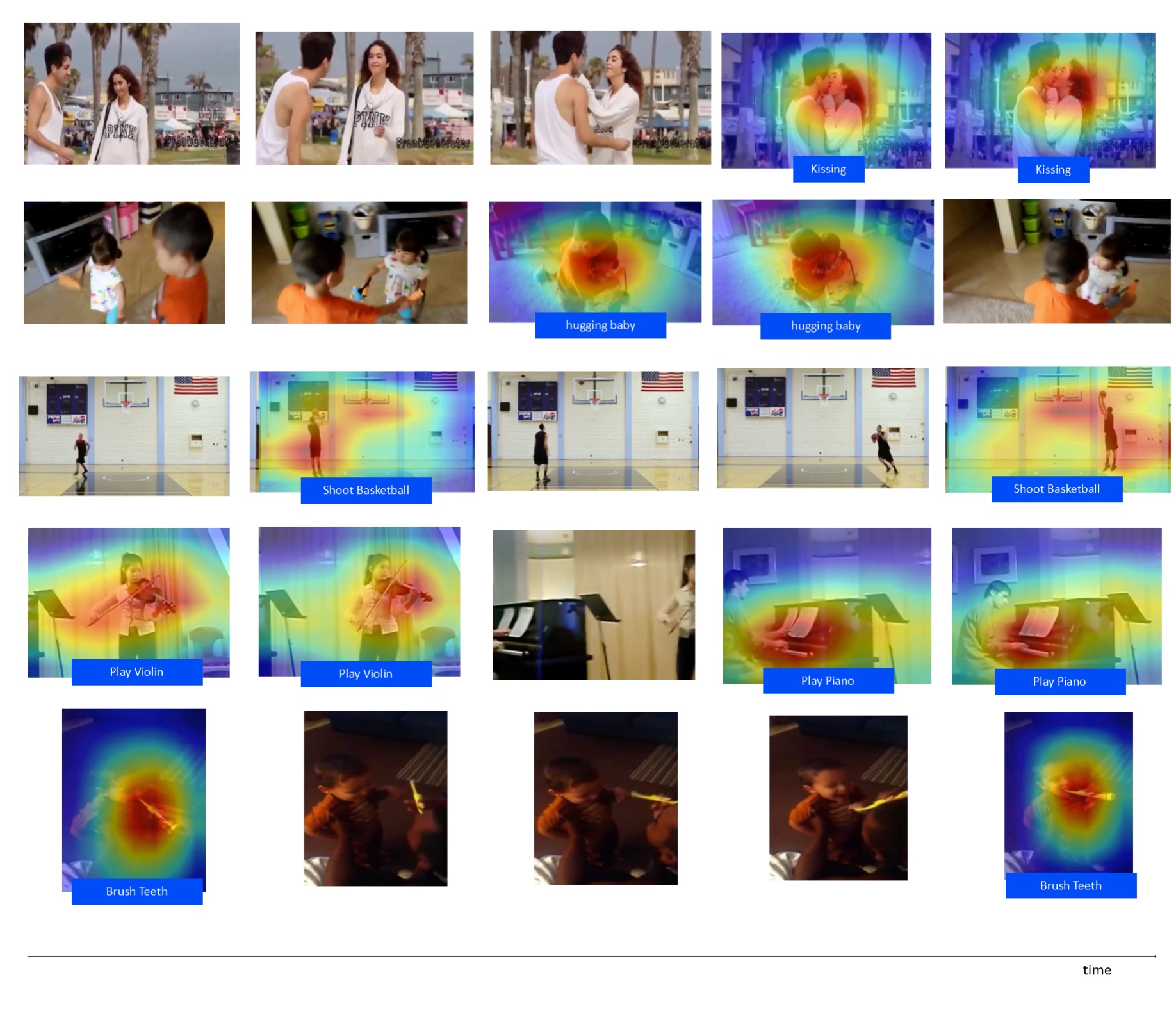}
\end{center}
    
   \caption{Visualization of spatial-temporal action event.}
\label{fig:visualization}
\end{figure*}

Figure \ref{fig:visualization} shows visualization examples for several videos. Here only clips with attention weights (i.e., $att$) that surpass the threshold (e.g., 0.5) will be marked as relevant segments with corresponding gradients being back-propagated to generate the spatio-temporal heatmap. As can be seen in Figure \ref{fig:visualization}, the GCF-Net nicely capture relevant clips in terms of temporal and spatial space. Despite the lack of temporal labels during training, the GCF-Network learns to judge whether it is a relevant or irrelevant clip unsupervisedly. The spatial location of an action is also nicely captured based on the back-propagation. The spatial pixel location of an action (e.g., the red region in Figure \ref{fig:visualization}) shows a good localization ability of the proposed network.

\section{Conclusion}

In recent years, most of the accuracy gains for video action recognition have come from the new design and exploration of 3D convolutional network architectures, which has very limited receptive field (e.g., merely 16 frames) and may not have enough representative power for a video level prediction.

This work aims to fill this research gap. In this work, we presented a light weight network to boost the
accuracy of existing clip-based action classifiers. It leverages two strategies: inter-clip fusion that explicitly models the inter-dependencies between video clips, and clip-wise importance that selects a relevant subset of clips for video level analysis. Experiments show that our GCF-Net yields large accuracy
gains on two action datasets with the cost of tiny increment in MFLOPs.

%
%
\bibliographystyle{splncs04}
\bibliography{egbib}
\end{document}